\documentclass{article}

\usepackage{algorithm}
\usepackage{algorithmic}
\usepackage[algo2e,linesnumbered,algoruled,boxed,lined]{algorithm2e}
\usepackage{arxiv}

\usepackage[utf8]{inputenc} 
\usepackage[T1]{fontenc}    
\usepackage{hyperref}       
\usepackage{url}            
\usepackage{booktabs}       
\usepackage{amsfonts}       
\usepackage{nicefrac}       
\usepackage{microtype}      
\usepackage{lipsum}		
\usepackage{graphicx}
\usepackage{doi}
\usepackage{amsmath}
\usepackage{amssymb}
\usepackage{soul}
\usepackage{multicol}
\usepackage{verbatim}
\usepackage{xcolor}
\usepackage{svg}

\title{PCV: A point cloud-based network verifier}

\author{ \href{https://www.arupsarker.com/}{\includegraphics[scale=0.06]{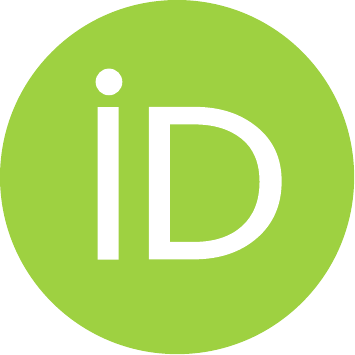}\hspace{1mm}Arup Kumar Sarker}\\
	Department of Computer Science\\
	University of Virginia\\
	Charlottesville, VA 22903 \\
	\texttt{djy8hg@virginia.edu} \\
	\And
	\href{https://www.linkedin.com/in/farzana-yasmin-ahmad-1b2235145/}{\includegraphics[scale=0.06]{orcid.pdf}\hspace{1mm}Farzana Yasmin Ahmad} \\
	Department of Computer Science\\
	University of Virginia\\
	Charlottesville, VA 22903 \\
	\texttt{fa7sa@virginia.edu} \\
    \And
    \href{https://matthewbdwyer.github.io/}{\includegraphics[scale=0.06]{orcid.pdf}\hspace{1mm}Matthew B. Dwyer} \\
	Department of Computer Science\\
	University of Virginia\\
	Charlottesville, VA 22903 \\
	\texttt{matthewbdwyer@virginia.edu} \\
}

\hypersetup{
pdftitle={PCV: A point cloud-based network verifier},
pdfsubject={A preprint Network Verifier},
pdfauthor={Arup Kumar Sarker, Farzana Yasmin Ahmad, Matthew B. Dwyer},
pdfkeywords={PCV, Point Cloud},
}

\begin{document}
\maketitle
\begin{abstract}
3D vision with real-time LiDAR-based point cloud data became a vital part of autonomous system research, especially perception and prediction modules use it for object classification, segmentation, and detection. Despite their success, point cloud-based network models are vulnerable to multiple adversarial attacks, where certain factor of changes in the validation set causes significant performance drop in well-trained networks. Most of the existing verifiers work perfectly on 2D convolution. Due to complex architecture, dimension of hyper-parameter, and 3D convolution, no verifiers can perform the basic layer-wise verification. It is difficult to conclude the robustness of a 3D vision model without performing the verification. Because there will be always corner cases and adversarial input that can compromise the model's effectiveness. 

In this project, we describe a point cloud-based network verifier that successfully deals state of the art 3D classifier PointNet and verifies the robustness by generating adversarial inputs. We have used extracted properties from the trained PointNet and changed certain factors for perturbation input. We calculate the impact on model accuracy versus property factor and can test PointNet networks' robustness against a small collection of perturbing input states resulting from adversarial attacks like the suggested hybrid reverse signed attack. The experimental results reveal that the resilience property of PointNet is affected by our hybrid reverse signed perturbation strategy.   
\end{abstract}


\begin{multicols}{2}

\section{Introduction}
\begin{figure*}[htbp]
\centering
\includegraphics[width= 0.6 \textwidth]{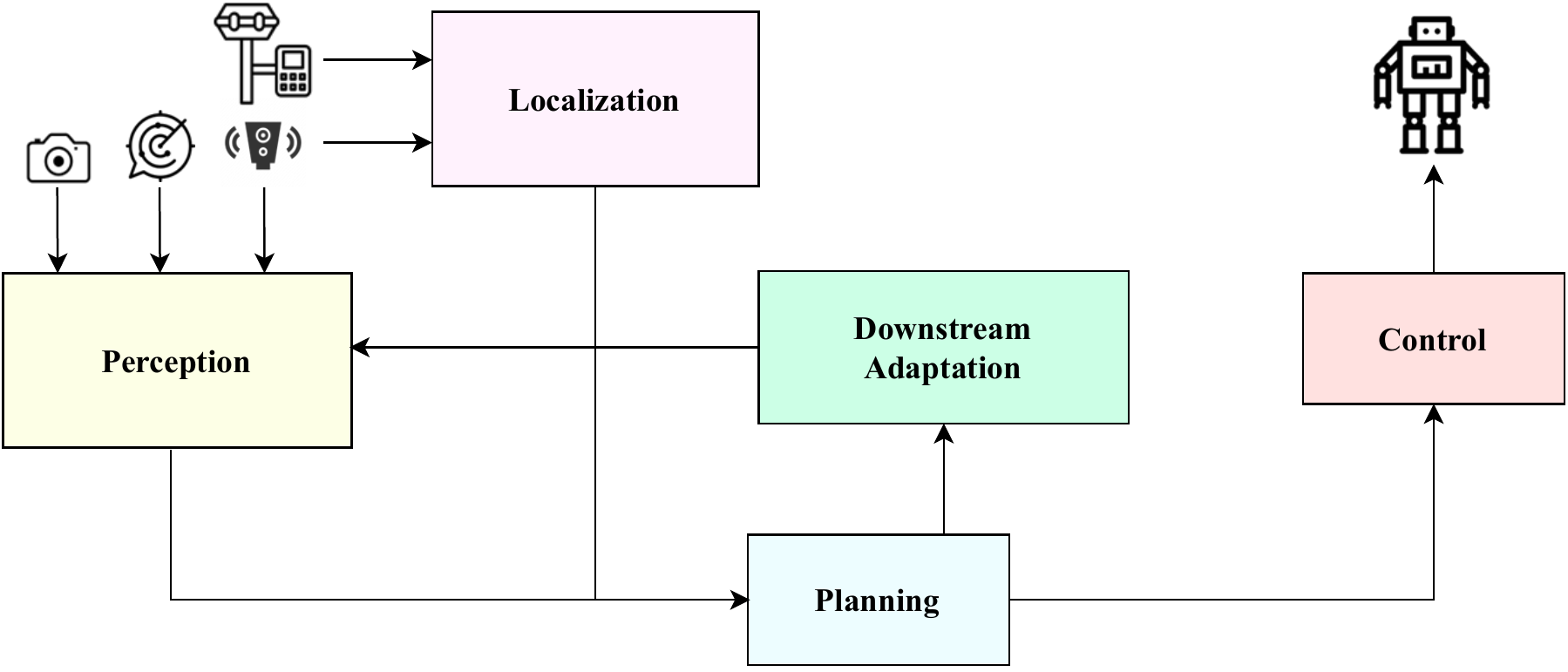}
\caption{High-Level flow of an Autonomous System}
\label{fig:architecture}
\end{figure*}

The point cloud is an important type of geometric data structure percept from the LiDAR in the autonomous system.  In \ref{fig:architecture}, the perception module process the data and detect 3D object. The planning tasks get the prediction results along with localization and send the driving policy to control so that control can send it to actuators. To implement weight sharing and other kernel optimizations in perception, typical convolutional architectures require extremely regular input data formats, such as picture grids and 3D voxels \cite{zhijian2019deep}. Because point clouds or meshes aren't in a standard format, most researchers convert them to 3D voxel grids or collections of images (e.g., views) before feeding them to a deep net architecture, which produces enormous data that obscures natural invariances. As a result, we concentrate on an alternative input representation for 3D geometry — point clouds – and call the deep nets PointNet\cite{charles20177pointnet}. Point clouds are easy to understand because they are basic and unified structures that avoid the combinatorial irregularities and complexities of meshes. 

\begin{figure*}[htbp]
\centering
\includegraphics[width=\textwidth]{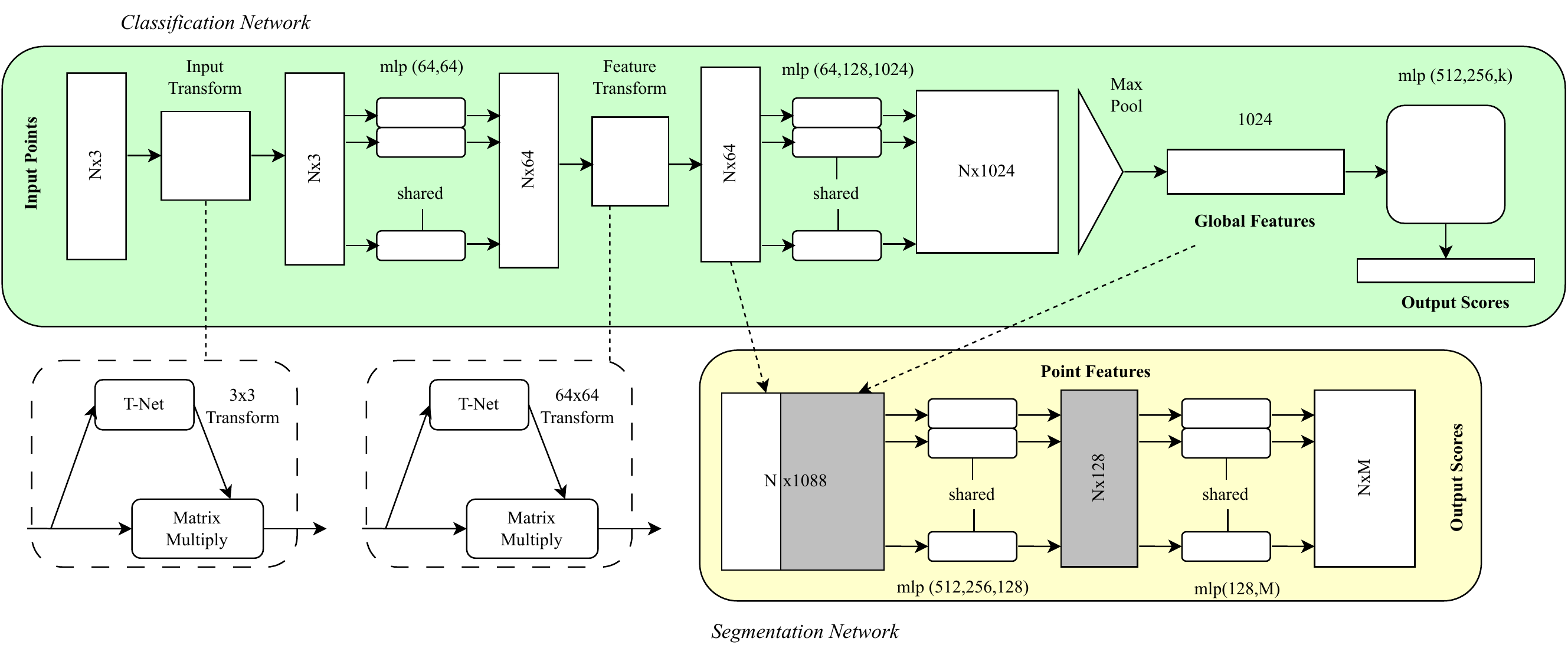}
\caption{Architecture of Pointnet \cite{charles20177pointnet}}
\label{fig:pointnet}
\end{figure*}

There could be adversarial assaults by placing dynamic noise on the input due to the widespread use of PointNet\cite{charles20177pointnet} and PointNet++\cite{qi2017pointnetplus} in perception modules of autonomous vehicles and robotics. Because tiny changes in the input could cause the network's accuracy and robustness properties to be violated at different layers (\ref{fig:pointnet}). Because of these concerns about employing  models in safety-critical applications due to their opacity, formally measuring the robustness of a trained PointNet is critical.

The majority of existing approaches focus on verifying the safety and robustness properties of feedforward neural networks (FNN) with the Rectified Linear Unit activation function (ReLU). There are several different approaches such as Mixed Integer Linear Programming (MILP) \cite{dutta2017output, kouvaros2018formal, lomuscio2017approach},  Satisfiability (SAT), and Satisfiability Modulo Theory (SMT) techniques \cite{ehlers2017formal, katz2017reluplex}, Optimization \cite{dvijotham2018dual, hein2017formal}, Geometric Reachability \cite{tran2019star, xiang2018output}  etc. Most of these works focus on 2D convolutional Neural Networks. All of these existing approaches use $L_0$ distance between two images. Their optimization-based approach computes a tight bound on the number of pixels that may be changed in an image without affecting the classification result of the network. These approaches do not fit with point cloud-based  data distribution. Overlapping points in the foreground bounding box, will create adversarial examples to improve the robustness of the network.

In this project, we implement a verifier for point cloud-based network model. Proposed method does not provide the robustness in terms of \textbf{number of points} that are allowed to be changed (\(L_0\) distance), \textbf{attacks} by disturbances, bounded with arbitrary linear constraints. These approaches are applied to CNN-based network verification. Even if we applied this to a point cloud-based network with a variety of measurements, it will be a novel approach. Rather, we add disturbance bounded with signed gradients and clipping into foreground bounding box points. In addition, we  add the reachability properties of a Region Pooling Layer for each validation set. It is a set-based analysis method by detecting the correctness and robustness properties. The representation can be used as a set of distorted points by an adversarial attack into the input domain. Target is to construct the reachable set of outputs from an adversarial attack that are used to reason about the overall robustness of the network. When a PoinNet-based network violates the robustness property, let’s say for detecting overlapping objects (e.g. pedestrian riding bi-cycle), an exact reachability scheme will construct a set of concrete adversarial examples. The contribution of this project is as follows.

\begin{itemize}
    \item PCV, a framework for point cloud network model based on efficient reachability analysis.
    \item Verification of correctness and robustness properties with the set of reachable objects that are considered adversarial objects.
    \item Implementation of PCV with reachability algorithm based on the over-approximate method
    \item Release the generated adversarial dataset for the future benchmark.
\end{itemize}

\begin{figure*}[htbp]
\centering
\includegraphics[width=0.5 \textwidth]{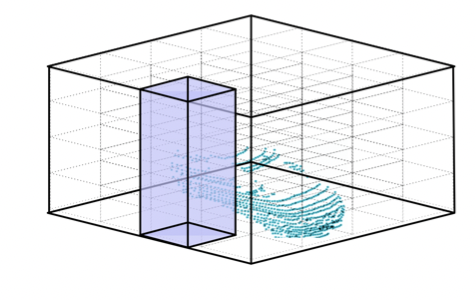}
\caption{Point clouds data of two overlapping objects which might be detected as a single object by PointNet}
\label{fig:overlap}
\end{figure*}


\section{Background}
\subsection{Point Clouds}
\begin{figure*}[htbp]
\centering
\includegraphics[width= 0.5 \textwidth]{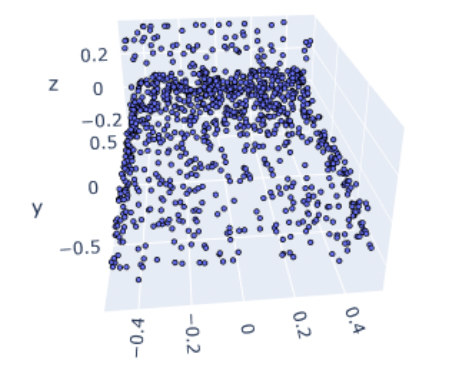}
\caption{An example of Point Clouds data from ModelNet dataset: bed}
\label{fig:bed}
\end{figure*}

Point clouds refer to a set of points in space and these points represent the 3D shape of the object. Cartesian coordinates$(X,Y,Z)$ are used to define the point position. Along with these coordinates point clouds data might include other information related to objects such as height, width, and length. An example of point clouds data from the ModelNet dataset is shown in figure \ref{fig:bed} Point clouds are usually generated from a 3D laser scanner and LiDAR (light detection and ranging) technology. There are several point clouds based dataset such as ModelNet\cite{wu20153d}, ShapeNet\cite{chang2015shapenet}, ScanNet\cite{dai2017scannet}, ApolloCar3D\cite{song2019apollocar3d}, PartNet\cite{mo2019partnet} etc.. Point cloud data is used in construction, highway planning, engineering, developing a self-driving car, augmented virtual reality, and housekeeping robots.

\begin{figure*}[htbp]
\centering
\includegraphics[width=0.6 \textwidth]{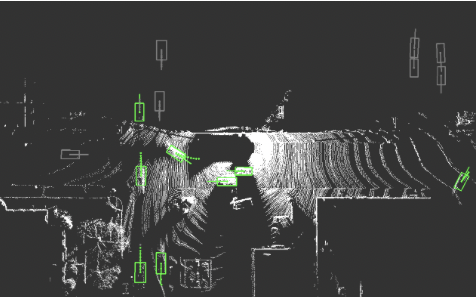}
\caption{This is a bird's eye view of LiDAR data. Objects inside the green boxes are cars. Here points are distorted. This opens a door for the adversarial attack}.
\label{fig:brdEye}
\end{figure*}

However, its irregular format of data is highly inconvenient to work with typical convolutional architecture as it requires a regular format of the input. To overcome this issue, researchers transform these point clouds or meshes into image grids or 3D voxels which are regular formats. Image grids or multi-view-based methods turn these unstructured point clouds data into 2D images, while volumetric-based method converts point clouds into 3D volumetric representations. Then the researchers can apply existing 2D or 3D convolutional networks which might cost the loss of information. On the other hand, point-based methods such as PointNet\cite{charles20177pointnet}, PointNet++\cite{qi2017pointnetplus} use direct point cloud data without any voxelization or projection. These methods do not cause any explicit information loss. PointNet can learn pointwise features and use the max pooling layer to gather global features. PointNet++ is a hierarchical network to detect fine geometric structures from the neighborhood of each point. However, a point-based method like PointNet can not detect overlapping objects which is a violation of the robustness property. For example, similar to Figure \ref{fig:overlap}, we have a point clouds data of a biker riding a bike. This PointNet-based network might not detect this point clouds data as human and bike separately. Then this example data can be considered as an adversarial example. Similarly, Figure \ref{fig:brdEye} shows that distortion of points in the point clouds can be a way of adversarial attack.

\subsection{Target model: PointNet}
PointNet\cite{charles20177pointnet} directly takes point clouds as input and classifies the entire input. Also, it can do per-point segmentation or labeling. The basic architecture includes points with three coordinates$(x,y,z)$. In figure \ref{fig:pointnet} we can see two networks: Classification and Segmentation networks. The classification network takes n points as input. There are transformations: input and feature transformation which are applied to these points respectively. This task is done by a mini network which is called T-net. An affine transformation matrix is determined by this T-net and then directly applied to the coordinates of the points. Then there is an aggregation of point features using the Max Pooling operation in order to make the model invariant to input permutation.  In the output, we get classification scores for $k$ classes. The segmentation network is an extension of the classification network, concatenating global and local features of the points to get per-point scores. PointNet is an ideal model to consider as the target model because of its simplicity and primitive design. With this, we can set a standard for verifying a model which has a common 3D vision task.

We face several difficulties while working with this model. We can not find any pertained model with onnx format. The original model is implemented in TensorFlow by the authors of the paper \cite{charles20177pointnet}. However, we have looked for the pytorch implementation of PointNet although there are some API conflicts for pytorch. We have to communicate with the authors to get the pytorch source. However, we have to spend a significant amount of time modifying the architecture to get it in the onnx format. The actual architecture is complicated. That is why we simply remove the segmentation network from our implementation of the verifier. 

\begin{figure*}[htbp]
\centering
\includegraphics[width=\textwidth]{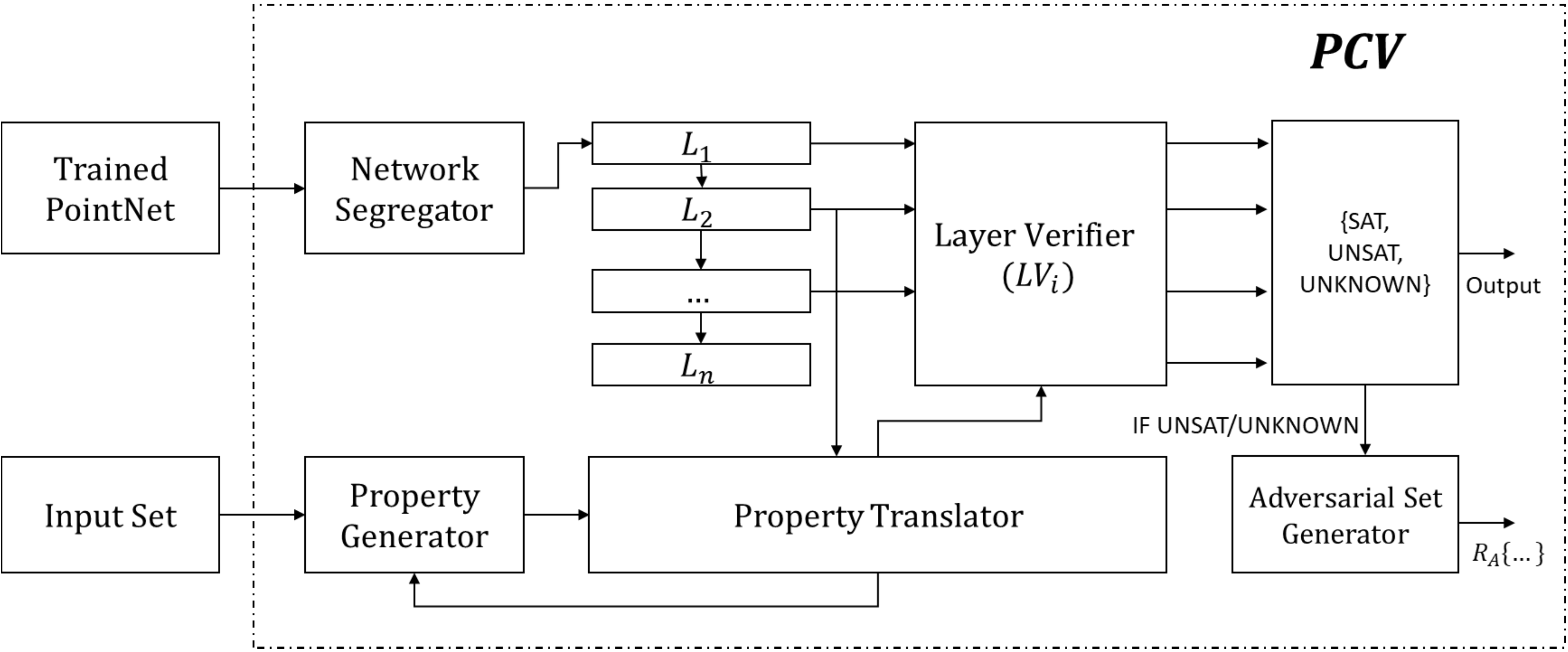}
\caption{Block Diagram of PCV}
\label{fig:arPCV}
\end{figure*}


\section{Problem Definition}

The robustness of a machine learning model defines how well a model is performing to classify or cluster the object. As a significant part of the training, data will come from multiple sources and we do not have full control over that. Therefore, it is very important to have robustness metrics for constant monitoring of the model, so that we can get a flag when it fails to result in a certain level of confidence. We are proposing a robustness framework consisting of the noise calibration module, target model, and robustness verifier. The final goal is to produce an adversarial set along with feature-wise impact analysis that is causing lower confidence in classification for the target model used.

Let's define the robustness properties with respect to a model \(f:\mathbb{R}^m \rightarrow \mathbb{R}^k\). Let \(\mathcal{X}\) denote the input space and \(\mathcal{Y}\) indicate the input and output spaces, respectively, and let  \(\mathcal{D}\)  be a random distribution over \(\mathcal{X}\)  from which input items are taken. Because we're dealing with a multi-label instance, it can be labeled with many values where \(\mathcal{Y} = {\{-1,+1\}}^k\). The learner is given a labeled sample \(S =\bigg((x_1, y_1),(x_2, y_2), . . . , (x_m, y_m)\bigg) \in (\mathcal{X} \times \mathcal{Y})^m\) with \(x_1, . . . , x_m\) drawn according to \(\mathcal{D},\)  and \(y_i = f(x_i) \) for all \(i \in [1,m],\  where\  f : X \rightarrow Y\) is the target labeling function. If we add a noise \(\mathcal{C}\) with \(\mathcal{X}\), then
\begin{align*}
    \bigg\{\big|\mathcal{X} - \mathcal{C}\big| &\leq \mathcal{T}\bigg\} \\
    r_1 &\leftarrow f(x) \\
    r_2 &\leftarrow f(x-c) \\
    \bigg\{ class(r_1) &= class(r_2)\bigg\}
\end{align*}
Here \(\mathcal{T}\) is the tripping point, after that the model will fail to predict and \(class(r_1) \neq class(r_2)\)

We consider the reachability of a \(PointNet \ P\) that consists of a series of \(layers \ L\) that include fully connected layers, max-pooling layers, average pooling layers, region pooling layers, and ReLU activation layers. 
Mathematically, we define a PointNet with \(p\) layers as  \(P = {L_i}, i = 1,2,3,...,p\). The reachability of the PointNet \(P\) is defined based on the concept of reachable sets corresponding to a linear set  \(I\).

\begin{figure*}[htbp]
\centering
\includegraphics[width= 0.5 \textwidth]{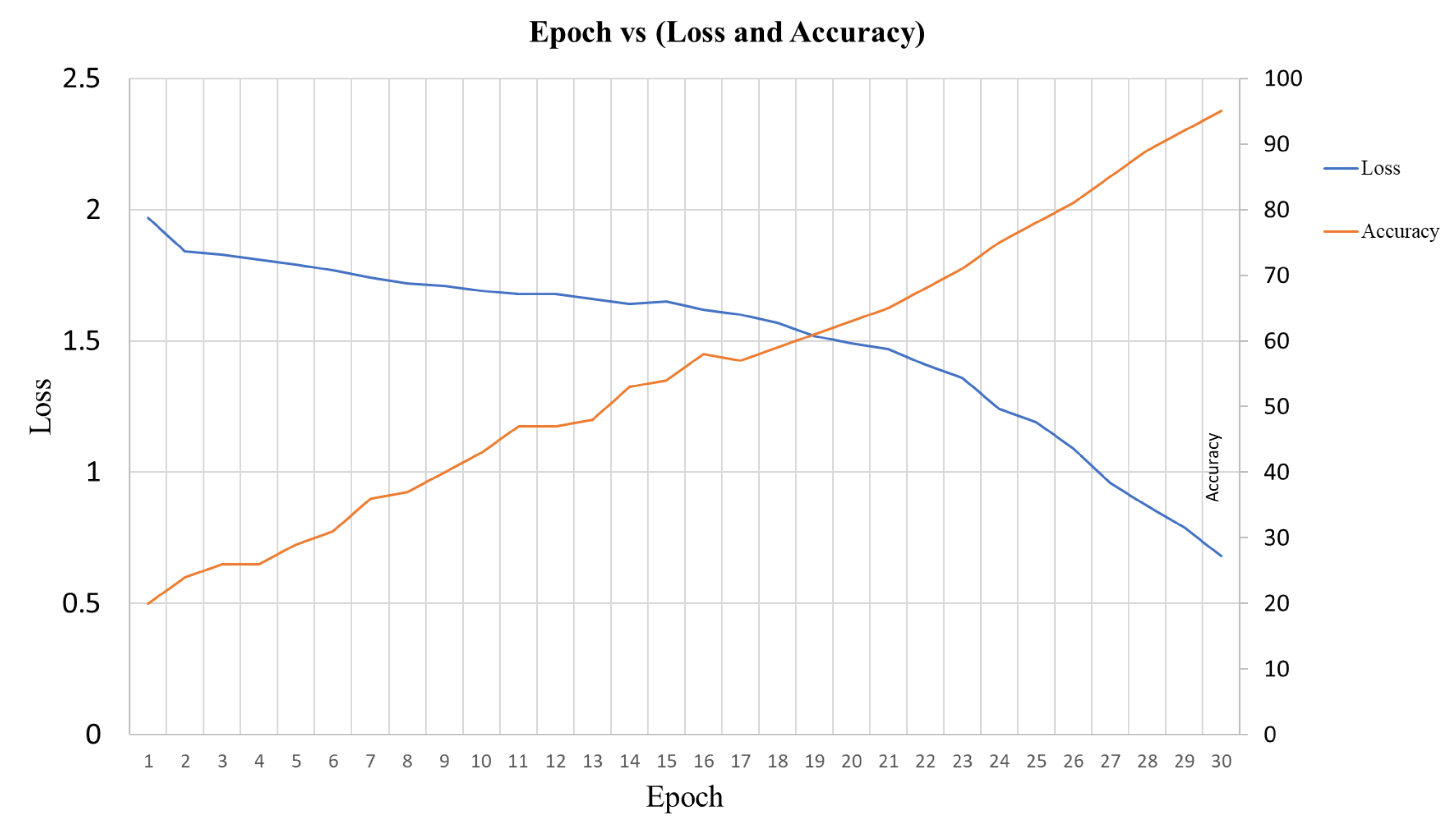}
\caption{Model Performance with original validation data}
\label{fig:point-net-accuracy}
\end{figure*}

\begin{align*}
R_{L_1} &\overset{\Delta}{=} {y_1|y_1 = L_1(x), x \in I}  \\
R_{L_2} &\overset{\Delta}{=} {y_2|y_2 = L_2(y_1), y_1 \in R_{L_1}}  \\
R_{L_3} &\overset{\Delta}{=} {y_3|y_3 = L_3(y_2), y_2 \in R_{L_2}}  \\
&\cdots \\
R_P = R_{L_p} &\overset{\Delta}{=} {y_p|y_p = L_{p}(y_{p-1}), y_{p-1} \in R_{L_{p-1}}} 
\end{align*}

The definition shows that the reachable set of the \(PointNet \ P\) can be constructed layer-by-layer. The core computation is constructing the reachable set of each layer \(L_i\) defined by a specific operation.

So, for reachable set \(R_{L_{1,2,...,p}}\), the final output \(y_p\) should hold the true value of postcondition i.e. the detected object with higher precision.

\section{PCV Overview}

In Figure \ref{fig:arPCV}, we have shown a basic block diagram of PCV. It will output the verification state of each layer whether the property is satisfied or not. At the same time, it will have an adversarial set generator. There will be multiple modules. \textbf{Property Generator} will generate the possible properties for the input layer of the model. As our target is to verify each layer, we will have a property translator from the second layer and onwards. \textbf{Property Translator} plays a vital role in collecting the state of each output of the previous layer from the \textbf{network segregator} and loops back to the property generator for the new property of the downstream layers. \textbf{Layer Verifier \((LV_i)\)}, gets properties for each layer to verify the output for the reachability set. It will run the algorithm of exact and over-approximate to ensure the correctness and robustness properties. Our primary plan is to take the onnx and PTH versions of PointNet and make a compatible verifier. There are two main components of PCV: Noise Model and Robustness Verifier.

\begin{figure*}[htbp]
\centering
\includegraphics[width=\textwidth]{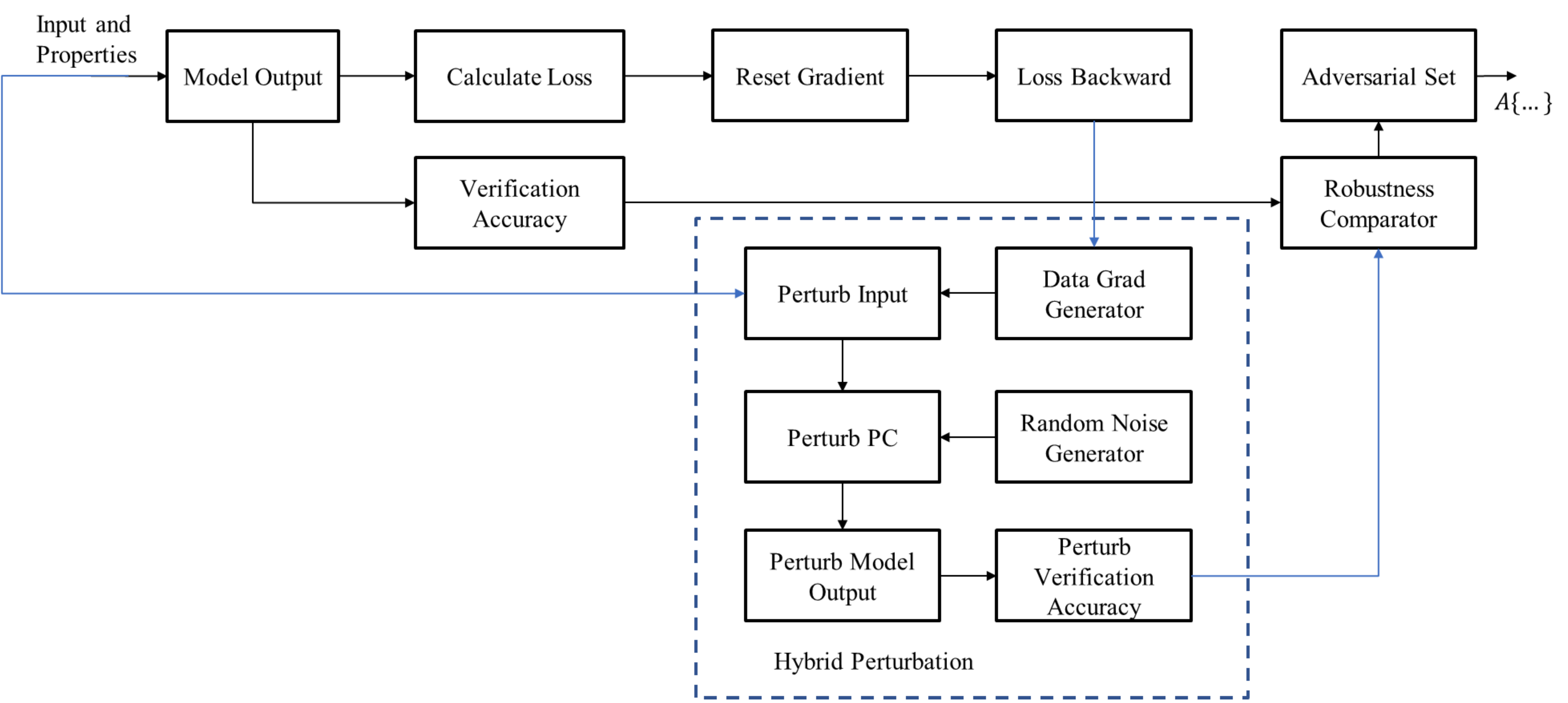}
\caption{Architecture of Robustness Verifier}
\label{fig:robustness_model}
\end{figure*}

\subsection{Noise Model}
The noise Model is the first basic component of PCV. How well a model can handle noisy data, define how robust the model is. The hybrid perturbation method adds noise into three stages. In the first stage, for \textbf{PCV noise generator}, we can define the function that creates the adversarial examples by perturbing the original inputs. The hybrid perturbation function takes three inputs, original clean input ($x$), `epsilon' is the element-wise perturbation amount $(\epsilon)$, and $data\_grad$ is the gradient of the loss w.r.t the input $(\nabla_{x} J(\mathbf{\theta}, \mathbf{x}, y))$ and $\eta$ is the noise function. After that same $\epsilon$ factor of Gaussian noise will be added to the perturbed input. Every real point will have neighbor noise points that have an impact during the convolution operation. Finally, a clipping operation will reset the value. The function then creates perturbed input as

\begin{align*}
perturbed\_input &= original\_input \\
&+ epsilon*sign(data\_grad) \\
&= x + \epsilon * sign(\nabla_{x} J(\mathbf{\theta}, \mathbf{x}, y)) \\ 
&= \eta(perturbed\_input)\\
&= \mathbf{Q}(perturbed\_input)
\end{align*}

In order to maintain the original range of the data, the perturbed input is clipped $\mathbf{Q}()$  to range $[0,1]$. This hybrid method generates more robust perturbation than FGSM\cite{goodfellow2014explaining}

\begin{figure*}[htbp]
\centering
\includegraphics[width=0.5 \textwidth]{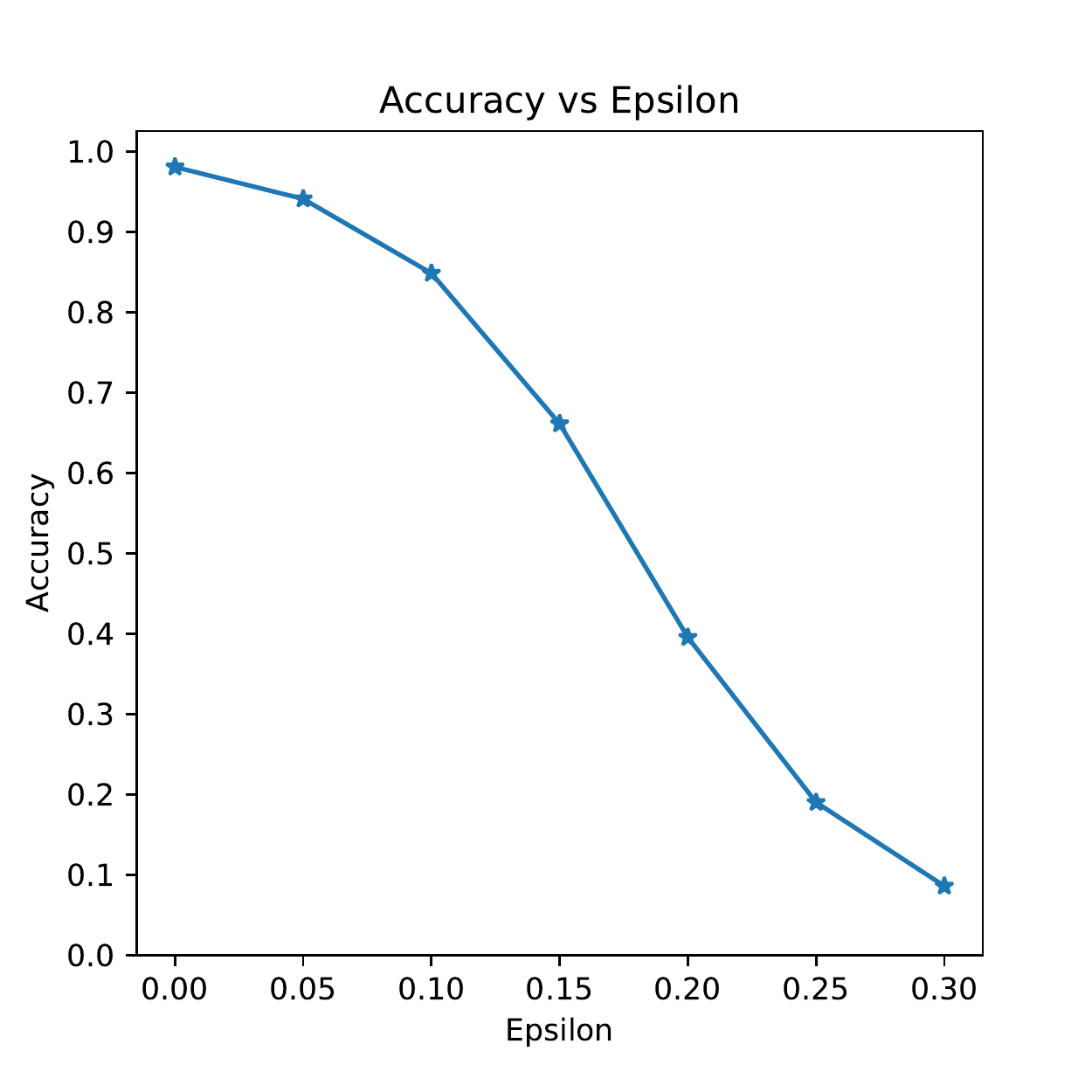}
\caption{Impact on accuracy due to increase of epsilon factors on PointNet}
\label{fig:accuracy_epsilon}
\end{figure*}

\subsection{Robustness Verifier}

The overall goal is to create the intended misclassification with the least amount of disruption to the input data. The enemy only cares about the output classification is incorrect, not the new classification. The concept is simple: instead of altering weights based on backpropagated gradients to minimize loss, the attack alters the input data to increase loss based on the same backpropagated gradients. In other words, the attack maximizes the loss by using the gradient of the loss in relation to the input data.

To clear the confusion, we state that all the perturbation process is done in the verification dataset. All models are trained with the original training set. There are multiple stages of the verification process that are handled by different modules. After the training process, we saved a trained model. The verification process starts by loading the trained model and generating the \textbf{Model Output} from the verification dataset. It has two parts. At first, we calculate the model \textbf{Verification Accuracy} and save it in a data structure. In parallel, we have to calculate the verification accuracy with perturb input. To do that, we \textbf{Calculate Loss} from the model output and \textbf{Reset Gradients}. Loss Backward is an important step for resetting the model state. In the next stage, hybrid perturbation is started. There are three different parameters: trained model, epsilon, and Data grad. In the beginning, we generate signs from the data grid and change the gradient steps by multiplying the input with epsilon. After that, add these two results and regenerate the \textbf{Perturb Input}. In addition, we add an epsilon factor of random noise to perturb input and exploit the property for the model. Finally, a clipped function is applied to perturb input and output of the final perturb verification dataset.

With that perturb input, we calculate the model output and generate \textbf{Perturb Verification Accuracy}. The robustness comparator will take two verification accuracy as input and results from the impact of noise into deviation of the accuracy. Based on the logic, if the verification results go down to the tipping point, we append that perturb dataset into \textbf{Adversarial Set}. We define the tipping point based on the standard of accuracy for a model. If it is an n-class classification problem, then below 50\% is the safest place to count on. Because there will be at least 2 classes for a multi-class model.   

\subsection{Verifier Algorithm}
We have presented the core ideas for the robustness analysis of PointNet. The adversarial set is constructed by every input with a certain factor of perturbation for which the accuracy of output drops. The full algorithm has two parts. We present the Algorithm-\ref{alg:cap} for step by step verification process of every single input dataset. It takes trained PointNet, epsilon, and validation dataset as input and returns the threshold tipping and adversarial data set. Algorithm-\ref{alg:perturb} is the process of generating noisy data into three stages. The robustness verification function uses it for generating perturbed input data. $HYBRID_P$ also takes three parameters as input: single point cloud frame, epsilon, and the gradients generated from the original set data. The epsilon factor is used for both steps to maximize loss and for generating Gaussian noise. These unique factors make the perturbation process more robust.  

\begin{algorithm}[H]
\caption{Robustness Verification in PointNet}\label{alg:cap}
\KwIn{$PNet,\  \epsilon,\  VSet$}
\KwOut{$\mathcal{T}\{...\},\  \mathcal{A}{\{...\}}$}
\begin{algorithmic}
\FORALL{$\epsilon$}
\FORALL{$VSet\_data from VSet$}
        \STATE {$data,\  target \leftarrow VSet\_data$} \\
        \STATE {$data.requires\_grad \leftarrow TRUE$} \\
        \STATE {$output = model(data) $}\\
        \STATE {$i\_pred \leftarrow output.max(1, keepdim=True)$} \\
        \IF{$i\_pred\  !=\  target$}
            \STATE{$I\_CC += f\_pred.item().sum()$}
        \ENDIF
        \STATE {$loss = F.nll\_loss(output, target) $} \\
        \STATE {$model.zero\_grad() $}\\
        \STATE {$loss.backward()$} \\
        \STATE {$data\_grad = data.grad.data $}\\
        \STATE {$p\_data = HYBRID\_P(data, \epsilon, data\_grad)$} \\
        \STATE {$output = model(p\_data) $}\\
        \STATE {$f\_pred \leftarrow output.max(1, keepdim=True)$} \\
        \IF{$i\_pred\  !=\  f\_pred$}
            \STATE {$\mathcal{A}\{\} \ \leftarrow VSet\_data$}
        \ELSE 
            \STATE{$F\_CC += f\_pred.item().sum()$}
        \ENDIF
        \STATE{$i\_acc = I\_CC/len(VSet)$}
        \STATE{$f\_acc = F\_CC/len(VSet)$}
        
        \IF{$f\_acc\  \leq\  i\_acc \times 50\%$}
            \STATE {$\mathcal{T}\{\} \ \leftarrow f\_acc$}
        \ENDIF
            
\ENDFOR
\ENDFOR
\RETURN {$\mathcal{T}\{...\},\  \mathcal{A}{\{...\}}$}
\end{algorithmic}
\end{algorithm}

\begin{algorithm}[H]
\caption{HYBRID\_P() Perturbation process into input data}\label{alg:perturb}
\KwIn{$x,\  \epsilon,\  data\_grad$}
\KwOut{$perturb\_x$}
\begin{algorithmic}
    \STATE {$s\_dg,\ \leftarrow data\_grad.sign()$}\\
    \STATE {$perturb\_x\  =\  x\  + \ \epsilon \times s\_dg$} \\
    \STATE {$perturb\_x\  =\  \eta(perturb\_x)$} \\
    \STATE {$perturb\_x\  =\  \mathcal{Q}(perturb\_x)$} \\
    \RETURN {$perturb\_x$}
\end{algorithmic}
\end{algorithm}

\section{Evaluation}
\label{evaluation}
\subsection{Experimental Setup}
We have used PointNet\cite{charles20177pointnet} as the target model to verify. There are two parts: classification and segmentation. In the current scope, the verification module works only for the classification part. We trained PointNet with the ModelNet dataset for 20 epochs with 10 batch sizes. It has 10 classes of point cloud data. Training and validation data sizes are 3991 and 908. After training with clean data, we saved the model into 'onnx' and 'pth' format. Our development environment configuration was Ubuntu 20.04 with Intel Core i9, 64GB physical memory, and Titan XP GPU with 12GB integrated memory. It took nearly 4 hours to train the whole model. For executing the robustness process with 7 different epsilon values, it took nearly 1hr 15 minutes.

\subsection{Experiment Results}
We re-sampled the input into $[64,\ 3,\ 1024]$, where 64 is the input dimension with 1024 sampled point clouds that have 3-dimensional$(x,y,z)$ coordinates. The validation function is the project's most important outcome. Each call to this test function runs the ModelNet test sets in full and reports the final accuracy. This function, however, also accepts an epsilon input. Because the validation function reports the accuracy of a model under attack from an adversary with strength $\epsilon$, this is the case. The function computes the gradient of the loss w.r.t. the input data ($data\_grad$), generate a perturbed image with "gradient\_sign\_p" ($perturbed\_data$), and then tests to see if the perturbed example is adversarial for each sample in the test set. The function saves and returns several successful adversarial samples in addition to verifying the model's correctness.

The final step in the process is to actually run the assault. For each epsilon value in the `epsilons' input, we conduct a whole test step. We also reserve the final accuracy and a few successful adversarial samples for each epsilon, which will be plotted in the following sections. As the epsilon value grows, the printed accuracies drop. Also, keep in mind that the $epsilon=0$ condition represents the initial test accuracy without any attacks. An increase of $epsilon$, means an increase in step size to maximize the loss. That means accuracy will be decreased. We will analyze the pattern of decreasing accuracy.  

\begin{figure*}[htbp]
\centering
\includegraphics[width= 0.7 \textwidth]{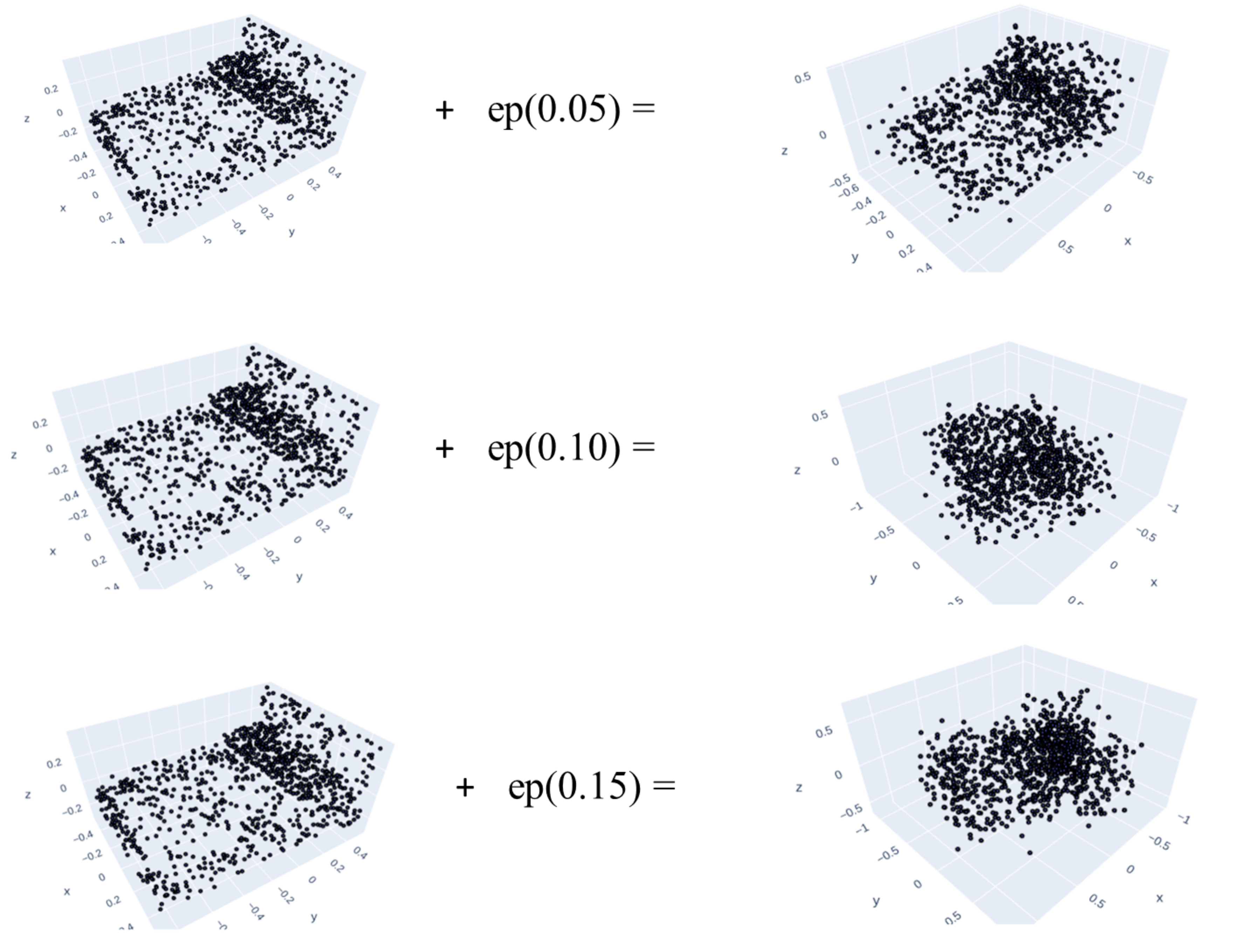}
\caption{Impact of noise on Point Clouds data}
\label{fig:noisy_data}
\end{figure*}

In Figure-\ref{fig:point-net-accuracy}, the validation results in accuracy for each class with 908 validation inputs. Although we developed a custom PointNet model, with only 10 epochs of training, validation shows good results. That is our base accuracy for verifying the robustness. We try to find the robustness threshold $\mathcal{T}$ when the accuracy  goes down less than $50\%$ for PointNet.

\subsection{Results Analysis}

For the verification of robustness, we plot the accuracy versus epsilon. The target is to find the tipping point where the model will fail to predict the desired classification. As previously stated, as epsilon increases, test accuracy should decrease. This is because greater epsilons indicate that we are taking a larger step in the direction of maximum loss. Even though the epsilon numbers are linearly spaced, the curve's trend is not linear. For instance, in Figure-\ref{fig:accuracy_epsilon},  while the accuracy at $epsilon=0.05$ is just roughly 5\% lower than $epsilon=0$, the accuracy at $epsilon=0.2$ is 20\% lower than $epsilon=0.15$. Also, the model's accuracy for a 10-class classifier between $epsilon=0.25$ and $epsilon=0.3$ hits random accuracy.

The test accuracy drops as epsilon grows, making the perturbations more visible. In reality, an attacker must weigh the tradeoff between accuracy degradation and perceptibility. At each epsilon value, we demonstrate some examples of effective adversarial examples in Figure-\ref{fig:noisy_data}. The epsilon value for each row of the figure is different. The first row shows the $epsilon=0$ samples, which are the clean photos that have not been altered. The initial classification $\rightarrow $ adversarial classification is shown in the title of each image. At $epsilon=0.15$, the perturbations become visible, and at $epsilon=0.3$, they are fairly visible. Despite the increased noise, humans are still capable of selecting the correct class in all circumstances. From Figure-\ref{fig:accuracy_epsilon}, when the value of $\epsilon = 0.20$ the accuracy falls to 43.1\%, which is below 50\%. Tipping point$\mathcal{T}$ for that specific adversarial input $(\epsilon = 0.20)$

We have submitted our project files into the github repository: {\color{blue}{\href{https://github.com/arupcsedu/PCV}{https://github.com/arupcsedu/PCV}}}


\subsection{Limitations and Future Work}
Due to complicated network architectures, we had to rewrite the PointNet and removed the segmentation part. PCV only works for the classification module. On top of PointNet, PointNet++ is currently state of art with the concept of hierarchical convolution, and PCV is not designed to verify that. It is the first verifier for 3D vision and we plan to support n-dimensional convolution in the future. In addition, any bounding box-based object detection models (e.g., PointRCNN\cite{Shi_2019_CVPR}), has more than 3D dimension input. Height$(h)$, Width$(w)$, Length$(l)$, Orientation$(\theta)$ are important dimension for real time 3D object detection. The KITTI dataset has that support. PCV only supports $(x,y,z)$ coordinates of data(e.g., ModelNet or ShapeNet). But by design, PCV can be customized with a more complex and robust dataset (e.g KITTI, Argoverse). In the next phase,  we plan to add verification of the segmentation module with PointNet. There are other 3D object detection models such as PointRCNN, PointNet++, VoxelNet\cite{zhou2018voxelnet} where Voxelnet has voxel 3D data. PCV can be extended for these models.  

\section{Related Work}


In our proposal, we mentioned using ImageStar\cite{tran2020verification}, a set-based framework for the verifier of CNN. The set-based representation and reachability algorithm is built in NNV. We could not use this verifier later due to some challenges. One of the major complications is its implementation which is available in MatLab and there are no full guidelines to use it given that it is a very new verifier. It was way complicated to convert this MatLab implementation to PyTorch implementation. Also, there is no source available for PointNet in MatLab.

\begin{figure*}[!htbp]
\centering
\includegraphics[width=\textwidth]{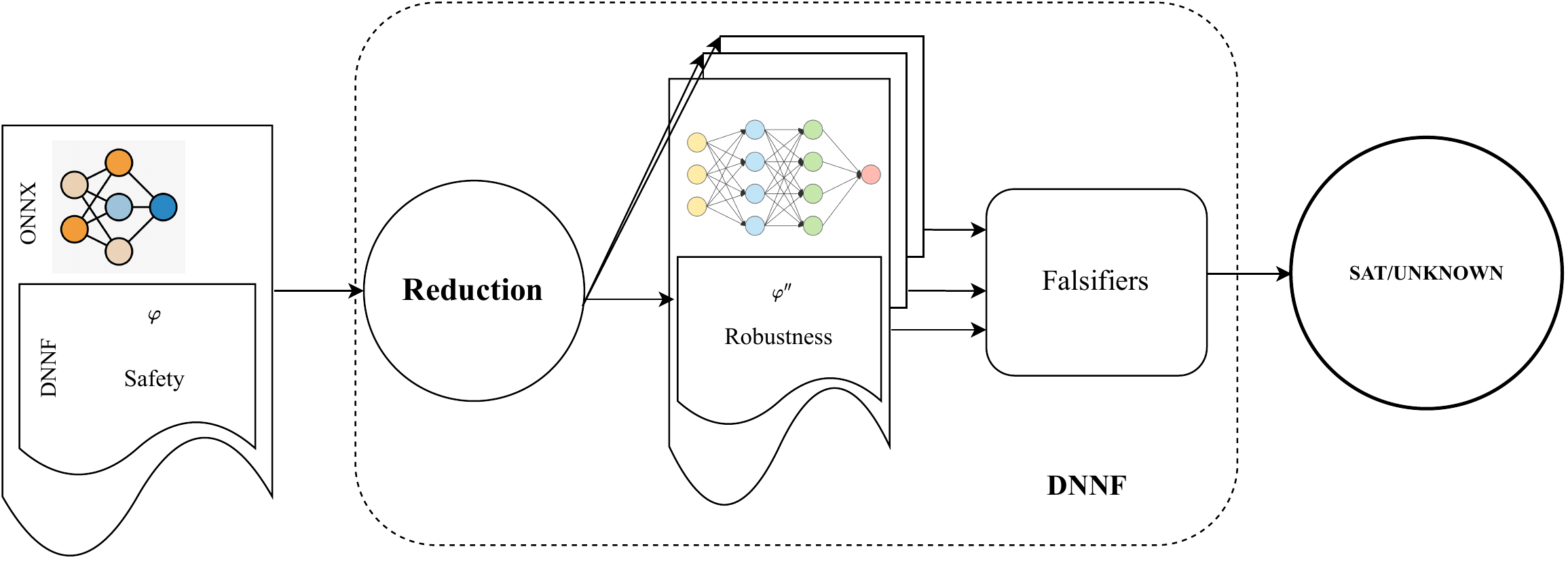}
\caption{Architecture of DNNF \cite{shriver2021reducing}}
\label{fig:DNNf}
\end{figure*}

\begin{figure*}[!htbp]
\centering
\includegraphics[width=\textwidth]{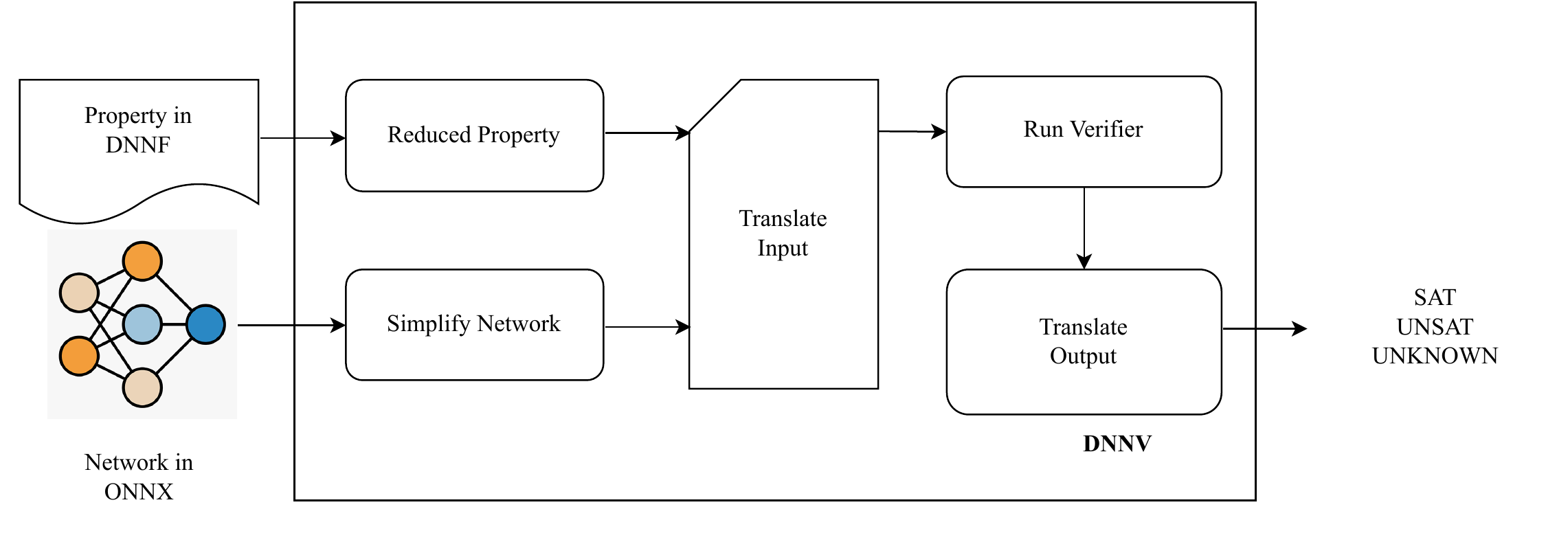}
\caption{Architecture of DNNV \cite{shriver2021dnnv}}
\label{fig:DNNV}
\end{figure*}

Other than ImageStar we try different existing verifiers. For this task we take the help of DNNV \cite{shriver2021dnnv}, an open-source tool supporting 13 neural network verifiers with extensive documentation. This DNNV framework makes the life of the DNN verifier researchers and developers easier by standardizing the inputs and output formats. For property, DSL is used. In figure \ref{fig:DNNV}, the architecture of this tool is explained. It takes the network in ONNX format as input, along with a property file written in Domain Specific Language DNNP and the name of the verifier. DNNV then translates the network format and property to be applicable for the target verifier and finally gives an output in a standard format. To verify our PointNet model we use two verifiers Neurify and Marabou. Neurify \cite{wang2018efficient} gives a tight output bound of a network for a given input range. It is also applicable to a larger network. Marabou\cite{katz2019marabou} is an extension of Reluplex \cite{katz2017reluplex} algorithm. Marabou provides native support for  fully connected and convolutional DNNs with arbitrary piecewise-linear activation functions. However, we fail to verify our Pointnet model in both verifiers. We got the error message \textit{``NotImplementedError: Non-2D convolutions are not supported"}. This is because DNNV currently only supports 2d convolutions and we have identified that at least one of the convolutions in our model is not 2d. This is because our point cloud  input has 3-dimensional coordinates as x,y,z. None of the verifiers will run because DNNV does not support the network.

We use another tool DNNF\cite{shriver2021reducing} which is based on falsification. A falsification is a complementary approach to verification. It tries to find out the violations of the properties. It can sometimes give output more quickly than verifiers when the property is false. DNNF can transform the correctness problem to equivalent sets of adversarial robustness problems using reduction\ref{fig:DNNf}. We decide to try this tool because DNNF supports a wide range of convolutional layers. However, we also fail here to run DNNF with the same error message as DNNV.

\section{Conclusions}
The correctness and Robustness properties of a Machine Learning model play a vital role to handle unknown and corner cases in large input domains. PCV is the first Point-Cloud Network verifier for the robustness of the network and successfully verifies the model properties by generating adversarial sets. Our proposed hybrid perturbation technique can exploit the properties of the model and compromise it. These adversarial sets create a state of the art examples for any future 3D model verification. Although we have played with basic properties to verify the network, there is still significant scope to verify other non-linear properties. In the next stages, we will extend this verifier for the more complex model in real-time 3D vision.

\bibliographystyle{abbrv}
\bibliography{main}  

\begin{thebibliography}{10}

\bibitem{chang2015shapenet}
A.~X. Chang, T.~Funkhouser, L.~Guibas, P.~Hanrahan, Q.~Huang, Z.~Li,
  S.~Savarese, M.~Savva, S.~Song, H.~Su, et~al.
\newblock Shapenet: An information-rich 3d model repository.
\newblock {\em arXiv preprint arXiv:1512.03012}, 2015.

\bibitem{qi2017pointnetplus}
H.~S. Charles R~Qi, Li~Yi and L.~J. Guibas.
\newblock Point-net++: Deep hierarchical feature learning on point sets in a
  metric space.
\newblock In {\em Proceedings of Advances in Neural Information Processing
  Systems (NIPS)}. NIPS, NIPS, 2017.

\bibitem{charles20177pointnet}
K.~M. Charles Ruizhongtai~Qi, Hao~Su and L.~J. Guibas.
\newblock Pointnet: Deep learning on point sets for 3d classification and
  segmentation.
\newblock In {\em Conference on Computer Vision and Pattern Recognition}. CVPR,
  CVPR, 2017.

\bibitem{dai2017scannet}
A.~Dai, A.~X. Chang, M.~Savva, M.~Halber, T.~Funkhouser, and M.~Nie{\ss}ner.
\newblock Scannet: Richly-annotated 3d reconstructions of indoor scenes.
\newblock In {\em Proceedings of the IEEE conference on computer vision and
  pattern recognition}, pages 5828--5839, 2017.

\bibitem{dutta2017output}
S.~Dutta, S.~Jha, S.~Sanakaranarayanan, and A.~Tiwari.
\newblock Output range analysis for deep neural networks.
\newblock {\em arXiv preprint arXiv:1709.09130}, 2017.

\bibitem{dvijotham2018dual}
K.~Dvijotham, R.~Stanforth, S.~Gowal, T.~A. Mann, and P.~Kohli.
\newblock A dual approach to scalable verification of deep networks.
\newblock In {\em UAI}, volume~1, page~3, 2018.

\bibitem{ehlers2017formal}
R.~Ehlers.
\newblock Formal verification of piece-wise linear feed-forward neural
  networks.
\newblock In {\em International Symposium on Automated Technology for
  Verification and Analysis}, pages 269--286. Springer, 2017.

\bibitem{goodfellow2014explaining}
I.~J. Goodfellow, J.~Shlens, and C.~Szegedy.
\newblock Explaining and harnessing adversarial examples.
\newblock {\em arXiv preprint arXiv:1412.6572}, 2014.

\bibitem{hein2017formal}
M.~Hein and M.~Andriushchenko.
\newblock Formal guarantees on the robustness of a classifier against
  adversarial manipulation.
\newblock {\em Advances in neural information processing systems}, 30, 2017.

\bibitem{katz2017reluplex}
G.~Katz, C.~Barrett, D.~L. Dill, K.~Julian, and M.~J. Kochenderfer.
\newblock Reluplex: An efficient smt solver for verifying deep neural networks.
\newblock In {\em International conference on computer aided verification},
  pages 97--117. Springer, 2017.

\bibitem{katz2019marabou}
G.~Katz, D.~A. Huang, D.~Ibeling, K.~Julian, C.~Lazarus, R.~Lim, P.~Shah,
  S.~Thakoor, H.~Wu, A.~Zelji{\'c}, et~al.
\newblock The marabou framework for verification and analysis of deep neural
  networks.
\newblock In {\em International Conference on Computer Aided Verification},
  pages 443--452. Springer, 2019.

\bibitem{kouvaros2018formal}
P.~Kouvaros and A.~Lomuscio.
\newblock Formal verification of cnn-based perception systems.
\newblock {\em arXiv preprint arXiv:1811.11373}, 2018.

\bibitem{lomuscio2017approach}
A.~Lomuscio and L.~Maganti.
\newblock An approach to reachability analysis for feed-forward relu neural
  networks.
\newblock {\em arXiv preprint arXiv:1706.07351}, 2017.

\bibitem{mo2019partnet}
K.~Mo, S.~Zhu, A.~X. Chang, L.~Yi, S.~Tripathi, L.~J. Guibas, and H.~Su.
\newblock Partnet: A large-scale benchmark for fine-grained and hierarchical
  part-level 3d object understanding.
\newblock In {\em Proceedings of the IEEE/CVF conference on computer vision and
  pattern recognition}, pages 909--918, 2019.

\bibitem{Shi_2019_CVPR}
S.~Shi, X.~Wang, and H.~Li.
\newblock Pointrcnn: 3d object proposal generation and detection from point
  cloud.
\newblock In {\em The IEEE Conference on Computer Vision and Pattern
  Recognition (CVPR)}, June 2019.

\bibitem{shriver2021dnnv}
D.~Shriver, S.~Elbaum, and M.~B. Dwyer.
\newblock Dnnv: A framework for deep neural network verification.
\newblock In {\em International Conference on Computer Aided Verification},
  pages 137--150. Springer, 2021.

\bibitem{shriver2021reducing}
D.~Shriver, S.~Elbaum, and M.~B. Dwyer.
\newblock Reducing dnn properties to enable falsification with adversarial
  attacks.
\newblock In {\em 2021 IEEE/ACM 43rd International Conference on Software
  Engineering (ICSE)}, pages 275--287. IEEE, 2021.

\bibitem{song2019apollocar3d}
X.~Song, P.~Wang, D.~Zhou, R.~Zhu, C.~Guan, Y.~Dai, H.~Su, H.~Li, and R.~Yang.
\newblock Apollocar3d: A large 3d car instance understanding benchmark for
  autonomous driving.
\newblock In {\em Proceedings of the IEEE/CVF Conference on Computer Vision and
  Pattern Recognition}, pages 5452--5462, 2019.

\bibitem{tran2020verification}
H.-D. Tran, S.~Bak, W.~Xiang, and T.~T. Johnson.
\newblock Verification of deep convolutional neural networks using imagestars.
\newblock In {\em International conference on computer aided verification},
  pages 18--42. Springer, 2020.

\bibitem{tran2019star}
H.-D. Tran, D.~Manzanas~Lopez, P.~Musau, X.~Yang, L.~V. Nguyen, W.~Xiang, and
  T.~T. Johnson.
\newblock Star-based reachability analysis of deep neural networks.
\newblock In {\em International symposium on formal methods}, pages 670--686.
  Springer, 2019.

\bibitem{wang2018efficient}
S.~Wang, K.~Pei, J.~Whitehouse, J.~Yang, and S.~Jana.
\newblock Efficient formal safety analysis of neural networks.
\newblock {\em Advances in Neural Information Processing Systems}, 31, 2018.

\bibitem{wu20153d}
Z.~Wu, S.~Song, A.~Khosla, F.~Yu, L.~Zhang, X.~Tang, and J.~Xiao.
\newblock 3d shapenets: A deep representation for volumetric shapes.
\newblock In {\em Proceedings of the IEEE conference on computer vision and
  pattern recognition}, pages 1912--1920, 2015.

\bibitem{xiang2018output}
W.~Xiang, H.-D. Tran, and T.~T. Johnson.
\newblock Output reachable set estimation and verification for multilayer
  neural networks.
\newblock {\em IEEE transactions on neural networks and learning systems},
  29(11):5777--5783, 2018.

\bibitem{zhijian2019deep}
Y.~L. Zhijian~Liu, Haotian~Tang and S.~Han.
\newblock Point-voxel cnn for efficient 3d deep learning.
\newblock In {\em 33rd Conference on Neural Information Processing Systems
  (NeurIPS 2019), Vancouver, Canada}. NeurIPS, NeurIPS, 2019.

\bibitem{zhou2018voxelnet}
Y.~Zhou and O.~Tuzel.
\newblock Voxelnet: End-to-end learning for point cloud based 3d object
  detection.
\newblock In {\em Proceedings of the IEEE conference on computer vision and
  pattern recognition}, pages 4490--4499, 2018.

\end{thebibliography}





\end{multicols}{}

\end{document}